\documentclass[a4paper, 10pt, conference]{template/ieeeconf}  
\usepackage{template2020/FG2020}
\usepackage{lipsum}
\usepackage{graphicx}
\usepackage[linesnumbered,ruled,vlined]{algorithm2e}

\usepackage{cite}
\usepackage{balance}

\FGfinalcopy 

\IEEEoverridecommandlockouts                              
\usepackage{amsmath} 
\usepackage{amssymb}  

\DeclareMathOperator*{\VIP}{VIP}

\newcommand{\norm}[1]{\left\lVert#1\right\rVert}

\newcommand*\rot{\rotatebox{90}}

\def\FGPaperID{121} 

\title{\LARGE \bf
Face Attributes as Cues for Deep Face Recognition Understanding
}

\author{\parbox{16cm}{\centering
    {\large Matheus Alves Diniz and William Robson Schwartz}\\
    {\normalsize
    Smart Sense Laboratory,\\Department of Computer Science,\\
    Federal University of Minas Gerais, Brazil\\
    {\textit{\{matheusad,william\}}}@dcc.ufmg.br
    }}
    \thanks{
      \scriptsize{
    Code available at https://github.com/MatheusAD95/fg2020-faceunderstanding
    }
    }
}

\definecolor{darkblue}{rgb}{0.1, 0.1, .6}

\begin{document}

\ifFGfinal
\thispagestyle{empty}
\pagestyle{empty}
\else
\author{Anonymous FG 2020 submission\\ \vspace{3em} Paper ID \FGPaperID \\}
\pagestyle{plain}
\fi
\maketitle

\begin{abstract}
  Deeply learned representations are the state-of-the-art descriptors for face
  recognition methods.
  These representations encode latent features that are difficult to explain,
  compromising the confidence and interpretability of their predictions.
  Most attempts to explain deep features are visualization techniques that are 
  often open to interpretation.
  Instead of relying only on visualizations, we use the outputs of hidden
  layers to predict face attributes.
  The obtained performance is an indicator of how well the attribute is
  implicitly learned in that layer of the network.
  Using a variable selection technique, we also analyze how these semantic
  concepts are distributed inside each layer, establishing the precise
  location of relevant neurons for each attribute.
  According to our experiments, gender, eyeglasses and hat usage can be
  predicted with over $96\%$ accuracy even when only a single neural output is
  used to predict each attribute.
  These performances are less than $3$ percentage points lower than the ones
  achieved by deep supervised face attribute networks.
  {
  In summary, our experiments show that, inside DCNNs optimized for face
  identification, there exists latent neurons encoding face attributes almost
  as accurately as DCNNs optimized for these attributes.
  }
\end{abstract}

\section{Introduction}
Deeply learned face representations are currently employed in many
state-of-the-art methods for face
recognition~\cite{zhao2019regularface,deng2019arcface}.
These representations are obtained from Deep Convolutional Neural Networks
(DCNNs) which are optimized to discriminate thousands of individuals.
The superiority of DCNN features is related to their
depth~\cite{krizhevsky2012imagenet}.
Between the input image and the final classification layer there are dozens of
hidden layers.
Each such layer learns a latent representation of the original image
based on the output of the previous layer.
The final hidden layer is connected to an output layer which is usually
supervised under the tasks of identification, verification or
both~\cite{schroff2015facenet}.

The meaning of each neuron in the output layer is defined by the user, and
thus can be precisely explained.
For instance, in the identification task, each neuron represents an identity,
and therefore should only be activated for inputs from the same individual.
However, the activation patterns of the output layer are specific to the
training set.
Thus it is preferred to use features from one of the hidden layers as the
representation for a generic face image.
While this choice is effective, neurons from these layers are latent features
that are not easy to interpret.
Therefore, it becomes challenging to explain the behavior of such recognition
systems.
Besides the improvement of user confidence, understanding the nature of DCNNs
features could also lead to new insights on how to increase the performance of
these networks.

\begin{figure}
\centering
  \includegraphics[width=0.48\textwidth]{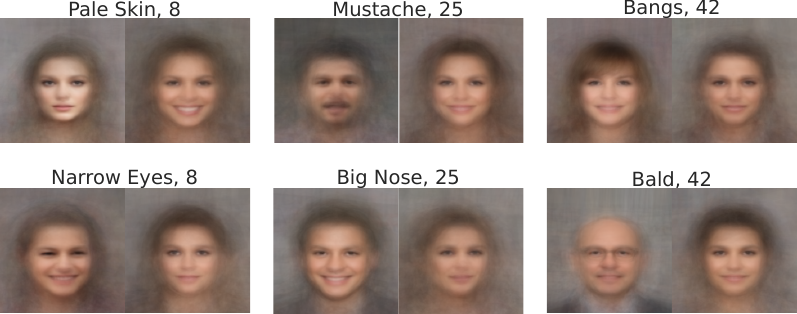}
  \caption{
    Average faces of the top-100 and bottom-100 activations of six selected
    neurons of a ResNet50 deep network that was optimized for face
    identification.
    Each neuron is identified by the attribute that it discriminates and its
    layer depth.
  }
  \label{fig:overview}
\end{figure}

Recently, several efforts have been dedicated to explaining the behavior of
DCNNs.
Visualization techniques are, perhaps, the most popular methods amidst such
efforts.
Instead of analyzing the raw output of a neuron, these methods create or
modify images to portray the patterns observed by that neuron.
For instance, some methods maximize the output of neuron with respect
to the input image~\cite{simonyan2013deep}.
When this neuron corresponds to some object class, the optimized input image
will contain patterns that resemble the object category.
By visualizing these images, users can diagnose the networks to both
understand failure cases as well as to gain intuition regarding the learning
process.

One inherent limitation of visualization techniques is their qualitative
nature.
This aspect may limit their applicability since their results are often open
to subjective interpretation.
Another limitation of visualization approaches is scalability.
Fine-grained analysis of individual feature-maps or neurons of the DCNNs is 
especially difficult to achieve.
A deep network may contain thousands of filters across its hidden layers, and
even more neurons.
Generating one image for each unit would be impractical due to time
constraints and the amount of generated images to be manually analyzed.

Discovering the meaning of each neuron in a network is an onerous task.
Besides the excessive number of neurons in a network, some of them may also
encode latent features that have no precise meaning for
humans~\cite{zhong2019exploring}.
To circumvent these adversities, we invert our goal.
Instead of observing a neuron to discover its meaning, we establish a few
concepts that are likely to be represented in the network and try to determine
which neurons encode them.
Particularly, we investigate the concepts learned by a deep face recognition
network.
Our hypothesis is that face attributes are encoded inside the DCNN, since
some of these attributes are relevant for recognition.
{
To verify this assumption, we train a traditional machine learning pipeline
to predict face attributes from the output of the hidden layers.
The accuracy of the predictor is used as an indicator of how well each
attribute is implicitly learned by the network.
To handle the high-dimensionality of the hidden layers, we employ
Partial~Least~Squares (PLS)~\cite{abdi2010partial} to create a discriminative
low-dimensional projection for each attribute. 
We then identify the most relevant neurons in the low-dimensional space
using the Variable Importance in Projection (VIP) technique.
Fig.~\ref{fig:overview}~shows a visualization of six neurons that were
identified with our approach.
For each neuron 
we show pair of images representing an average face of the inputs that
presented the highest and lowest activations.
}

According to our experimental results, a single neuron in the recognition
network can achieve, for some attributes, an accuracy exceeding $96\%$, which
is less than $3$ percentage points (p.p.) lower than a fully supervised face
attribute DCNN.
{
Such extraordinary performance demonstrates that face recognition DCNNs are
able to implicitly learn these semantic concepts without any supervision.
To the best of our knowledge, this is the first work that was able to provide a
thorough analysis of DCNNs using latent features as input for classifiers.
}

\section{Related Works}
The performance of automatic face recognition soared in the last years.
Deeply learned features coupled with immense amounts of
data~\cite{kemelmacher2016megaface,cao2018vggface2} are the main source of
this improvement.
State-of-the-art convolutional neural networks are trained to classify
thousands of identities, resulting in latent representations inside the
network that are very robust and
discriminative~\cite{sun2014deep,deng2017marginal, deng2019arcface}.
Deep networks can also be used to learn face attributes such as
gender, age or hair color~\cite{liu2015deep}.
Besides the direct application in retrieval, these attributes are also
beneficial to other methods such as verification, identification and
localization~\cite{ranjan2017all}.
In this work, we are not concerned with outperforming benchmarks on any of
these tasks.
Instead, we want to understand the learning of process of deep face
recognition networks, specifically of face attributes encoding inside the
network.

Many attempts to understand DCNNs consist of visualizing the learned filters
of intermediate layers.
Some approaches attach a
deconvolutional~\cite{zeiler2014visualizing,zhong2019exploring} or
up-convolutional~\cite{dosovitskiy2016inverting} network to a layer of the
DCNN to revert its feature maps back to the pixel space.
Other approaches rely on backpropagation to visualize images that are good
representations of some features of the
network~\cite{simonyan2013deep, yosinski2015understanding,
mahendran2016visualizing, nguyen2016multifaceted}.
This is achieved by optimizing a neural activation pattern with respect to the
input image, adjusting the input to match or maximize the desired pattern.
However, regardless of the method, it is unfeasible to visualize all filters
inside a network, specially with the growing size of network architectures.
Furthermore, the conclusions obtained by these approaches may be biased since
assigning semantic meaning to the obtained images is a subjective task.
Human operators are able to find interpretable meaning for network patterns
even when they are changed to a random basis~\cite{szegedy2013intriguing}.

The aforementioned limitations can be mitigated through a more quantitative methodology.
Some methods are able to visualize the region in the input image that is
more relevant for the classification output~\cite{zhou2014object,
zhou2016learning, selvaraju2017grad}.
In an image classification context, these regions represent the localization
of the class object in the image.
When compared to other methods that are supervised for localization, these
methods still have a mildly competitive performance, demonstrating that
DCNNs supervised only for classification encode information relevant for
object detection.

Deep face recognition contains few particularities that distinguishes it
from other DCNNs.
Traditionally, the supervised output of the training phase is also the desired
output of testing phase.
Face recognition networks, on the other hand, act as a feature
extractor at this second phase.
The supervised layer is removed, and the features of a hidden layer are
used along with another learning method or distance metric.
Thus, the output layer analysis is not as useful for face recognition
networks.

The attempts to understand face recognition networks are concentrated on
the analysis of the hidden neurons.
Examination of neuron activations revealed that even a single neuron is
highly discriminative for some face attributes or even specific
identities~\cite{sun2015deeply, ferrari2019discovering, liu2015deep}.
Furthermore, the neural activation pattern also encodes the quality of the
input image.
A feature with low L2-norm strongly indicates a hard to recognize input face
in a DCNN supervised for face identification~\cite{ranjan2017l2}.
Visualization techniques have limited usage in hidden layer analysis because
latent features do not necessarily encode any interpretable meaning for
humans.
However, a few interpretable feature maps can be found by detecting high
activation patterns for visually similar images~\cite{zhong2019exploring}.

Our approach engages the face recognition problem in a different manner.
Instead of visualizing or measuring activation patterns in the network to
define their meaning, we start from a set of attributes that are likely to be
encoded in the network and then use traditional machine learning techniques to
determine which features encode them.
Specifically, we train and validate an attribute classifier with the outputs
of each layer in the network, and use its weights and performance to infer how
the recognition behaves.
The usage of classifiers using middle-level representations of the network has
been previously studied in other contexts such as
transfer~learning~\cite{zeiler2014visualizing},
hypernets~\cite{jordao2018latent, kong2016hypernet}, early
predictions~\cite{shafiee2018efficient} and network
pruning~\cite{jordao2018pruning}.

\section{Proposed Approach}
We present a method to analyze deep face recognition networks.
For each layer of the network, we attach and train a face attribute classifier
and use its accuracy as an indicator of how well a face attribute is encoded
in that layer.
We believe that using the classifier performance as our indicator, diminishes
the effect of the human bias that is present in some visualization techniques.

One challenging aspect of our methodology is the high dimensionality of the
latent feature maps, which can exceed $800{,}000$ dimensions.
For such high-dimensional spaces, classical learning algorithms are
impractical to train due to small sample size, slow convergence or expensive
computation.
There is also a concern not to interfere too much with the network
representation.
For instance, using a convolutional sub-network as our classifier could add
semantic information that was not present in the original feature map.

In this context, dimensionality reduction techniques are suitable candidates
to overcome these problems.
Specifically, PLS~\cite{abdi2010partial} was previously shown to be capable of
dealing with dimensionalities as high as $170,000$~\cite{schwartz2009human}.
In its essence, PLS decomposes the flattened network features, $X$, and the
attributes matrix, $Y$ into
\begin{align}
  X & = TP^T + E\\
  Y & = UQ^T + F
\end{align}
where $T$ and $U$ are the extracted low-dimensional factors, $Q$ and $P$
contain the loadings, and $E$ and $F$ represent the residuals.
Algorithm~\ref{alg:nipals} shows how non-linear iterative partial least
squares (NIPALS) solves these equations.
NIPALS is able to retain discriminative information by exchanging the scores
$u_i$ and $t_i$ while factorizing $X$~and~$Y$~\cite{geladi1986partial}.
We execute the algorithm separately for each column of $Y$ to avoid
interference between the attributes on the projected space.

\begin{algorithm}[b]
  \SetKwInOut{Input}{input}
  \SetKwInOut{Output}{output}
  \SetKwRepeat{Do}{do}{while}
  \Input{X $\in \mathbb{R}^{n \times m}$, y $\in \mathbb{R}^{n \times 1}$,
        k $\in \mathbb{I}$}
  \Output{W* $\in \mathbb{R}^{m \times k}$}
  \BlankLine
  $E = X$\;
  $f = y$\;
  \For{$i\leftarrow 1$ \KwTo $k$}{
    randomly initialize $u_i \in \mathbb{R}^{n \times 1}$\;
    \Repeat{$w_i$ reaches\ convergence}{
      $w_i = \frac{E^Tu_i}{\norm{E^Tu_i}}$\;
      $t_i = Ew_i$\;
      $q_i = \frac{f^Tt_i}{\norm{f^Tt_i}}$\;
      $u_i = fq_i$\;
    }
    $p_i = E^Tt_i$\;
    $E = E - t_ip_i^T$\;
    $f = f - t_iq_i^T$\; 
  }
  $W^* = W(PW)^{-1}$\;
  \caption{NIPALS}
  \label{alg:nipals}
\end{algorithm}

\begin{figure}
  \centering
  \includegraphics[width=1\columnwidth]{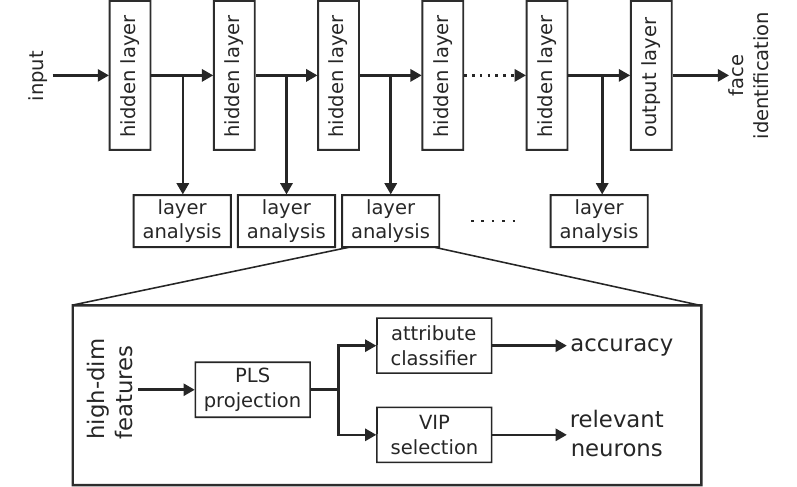}
  \caption{
    For each attribute, we learn a discriminative low-dimensional projection
    using the high-dimensional output of the deep network layer.
    The low-dimensional projection is used to measure how well the layer
    encodes an attribute, and to identify the most relevant neurons.
    The process is repeated for each convolutional layer of the DCNN.
  }
  \label{fig:proposed_approach}
\end{figure}

We also provide a fine-grained analysis of each layer of the network to
identify filters and neurons that encode relevant information for attribute
prediction.
To this end, we employ VIP~\cite{mehmood2012review} to measure how much each
feature, or neural output, of original high-dimensional space contributes to
the discriminative low-dimensional projection.\\
The VIP score of the $j$th neuron is determined by
\begin{align}
  \VIP(j) & = \sqrt{m\sum_{i=1}^{k}{\frac{SS_i\frac{W_{ij}}{\norm{W_i}^2}}{\sum SS_i}}},
\end{align}
where $m$ and $k$ are the original and the projection dimensionalities
respectively, and $SS_i$ is the sum of squares explained by the $i$th
component, which can be alternatively expressed as $q_i^2t_i^Tt_i$.

{
Fig.~\ref{fig:proposed_approach} summarizes our approach.
For each layer in the network, we learn a low-dimensional projection using
PLS, which is then used to train a simple classifier for each attribute.
This produces a performance curve for each attribute over the network depth,
showing how and where the network learns each attribute.
We also apply VIP technique over the PLS projection to determine the relevant
neurons of each layer.
Furthermore, we also propose to obtain the importance of each filter by
averaging the VIP score of its neurons.
The proportion of relevant neurons/filters at each layer reveals whether
the attributes information is concentrated in a few units or entangled
throughout the entire layer.
Finally, we also measure the performance of the classifier when only the
relevant neurons/filters are available as input so that we can quantify the
efficiency of the VIP selection technique.
}

\begin{figure*}[!ht]
  \includegraphics[width=\textwidth]{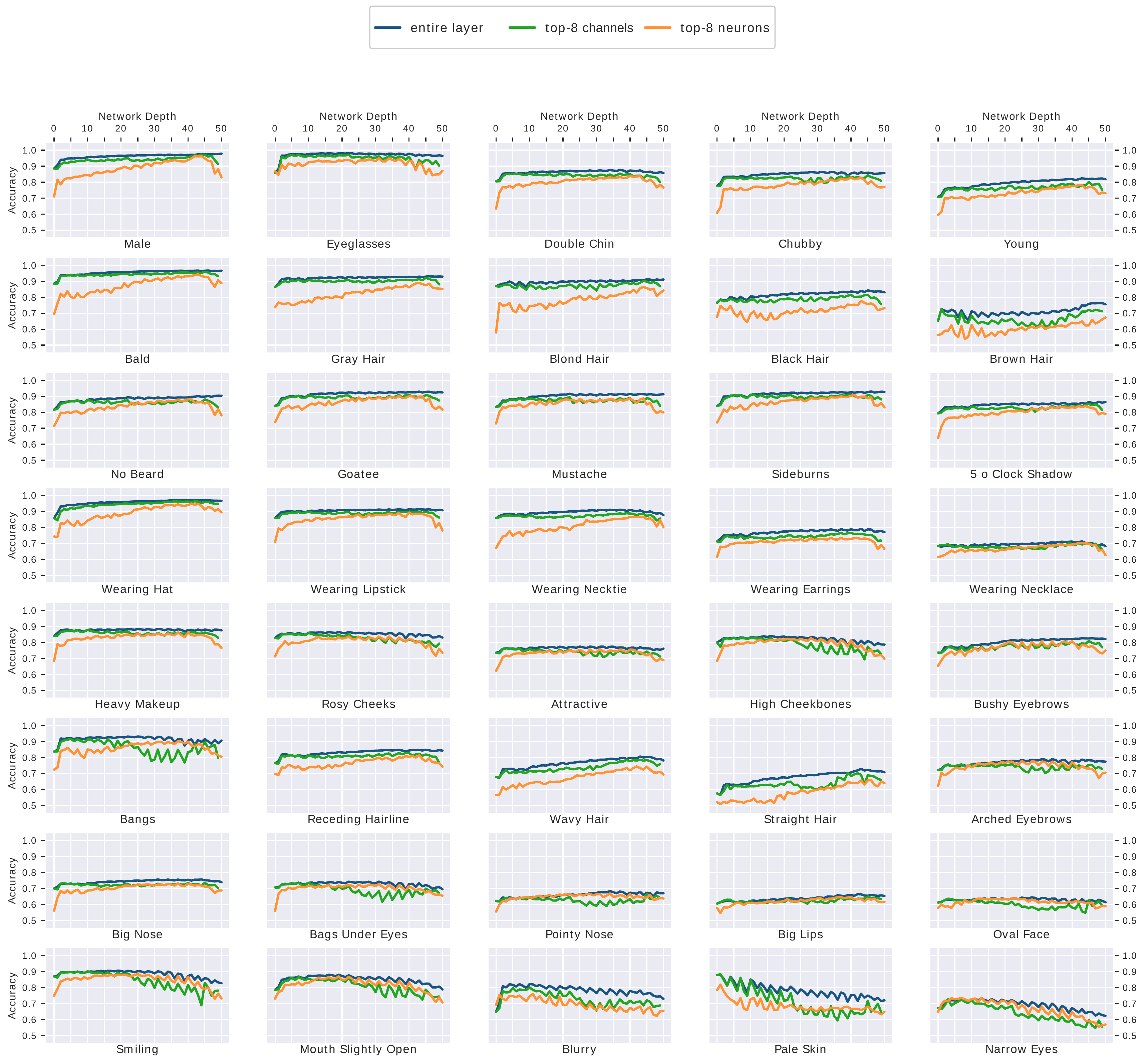} 
  \caption{Attribute accuracy over the network depth. The blue curve measures
  the accuracy when the entire layer output is used to predict the
  attribute. The green and orange curves measure the accuracy when only the
  best eight filters or neurons, respectively, are used for prediction.}
  \label{fig:pls_layerwise}
\end{figure*}

\section{Experimental Results}

Our face recognition networks are trained on the
\mbox{VGGFace2}~\cite{cao2018vggface2} dataset, using the
LFW~\cite{huang2008labeled} dataset as validation.
We followed the standard training protocol~\cite{cao2018vggface2}, using
uniform sampling of the identities, randomly cropping a $224 \times 224$
region from the resized image of which the shorter side is $256$ pixels,
and transforming the image to grayscale with a $20\%$ chance.
We adopt ResNet50~\cite{he2016deep} as our architecture because it provides a
strong performance on recognition tasks while still being fairly linear,
which allows us to represent the network performance as a line chart.

We use $40$ binary attributes from the CelebA~\cite{liu2015deep} dataset to
help us understand the recognition network.
Unless stated otherwise, all experiments were performed on the training and
validation splits of this dataset.
Even though PLS is more efficient than other classic learning
methods~\cite{schwartz2009human}, it still cannot load the entire training set
on memory. For each attribute, we sample $2{,}048$ faces for training, and $6{,}144$ for
validation from the original training and validation splits combined.
For all attributes, half the samples are positive labels, and thus, there
are no biases in our initial analysis.

We trained a PLS model to learn a projection of $8$ components for each
attribute and layer output.
Each projection was used to train a QDA classifier, which we then evaluate on
a validation set.
The dark blue curve in Fig.~\ref{fig:pls_layerwise} shows the achieved accuracy
for each attribute as a function of the DCNN feature depth averaged over six
experiments.
Each experiment is a different random optimization of the ResNet50
architecture, but the training and validation splits for the face attributes
are maintained throughout the experiments.

The general behavior of the curves is either an early plateau or a modest
slope after the first few layers.
This indicates that these shallow features are discriminative enough, or at
least competitive against deeper features, for most of the evaluated attributes.
Previous works suggest that shallow features are only capable of capturing
low level information such as textures and
edges~\cite{zhong2019exploring,mahendran2016visualizing}.
Human operators may visualize these features and assign no meaning to them,
while their high accuracy indicates that they already encode relevant semantic
information.
A contrasting hypothesis is that these features are so high-dimensional that
they are discriminative of any simple attribute, i.e., a bag of any set of
shallow features would also be discriminative of these attributes.
However, it remains unclear why some semantically equivalent attributes
have such vastly different performances, e.g., gray hair is more accurately
predicted than brown hair.

Interestingly, some soft-biometric attributes such as \textit{Wearing Hat} and
\textit{Eyeglasses} have exceptional accuracy.
This means that the network encodes such attributes, and that they are
relevant for the face recognition task.
In a surveillance scenario, these attributes could compromise the performance
of the recognition system since, in this setting, the system cannot rely on the
individual wearing his usual attire.
However, in the context of the VGGFace2 dataset optimization, it is intuitive
that each identity has some accessories of preference, and thus, the DCNN
uses this information as a cue for recognition.

\begin{figure}[t]
  \includegraphics[width=0.49\textwidth]{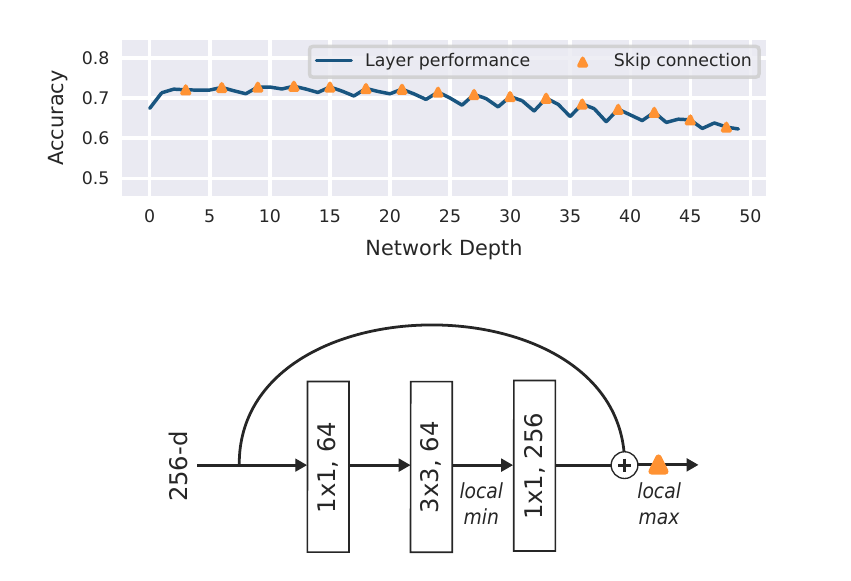} 
  \caption{Top: entire layer performance of the attribute Narrow Eyes. Bottom:
  residual block from the ResNet50 architecture. The performance curve
  exhibits a sawtooth pattern due to skip connections and
  bottlenecks present in the residual block.
  }
  \label{fig:narrow_eyes}
\end{figure}

Another interesting aspect is observed in the last row of
Fig.~\ref{fig:pls_layerwise}.
The performance of these attributes decreases as the network gets deeper.
These attributes encode intra-class variations of an identity, and thus, are
not discriminative for recognition.
For instance, the pale skin attribute indicates a variation of increased
brightness in a person with fair skin tone.
Thus, the accuracy drop indicates that the deeper representations in the network
are more robust to variations in brightness, blur or mouth pose.

The curves of the last row also exhibit a distinct sawtooth pattern.
In fact, all other curves display, albeit more subtly, a similar pattern.
Fig.~\ref{fig:narrow_eyes} shows a more detailed visualization of the Narrow
Eyes accuracy, which reveals that the sawtooth aspect is closely related to
the ResNet50 design.
This architecture is constructed with the concatenation of residual
blocks, which contain three convolutional layers each.
The first two layers of the block create a compact projection of the previous
block, reducing the number of filters of the representation, which is then
restored at the third layer to be combined with the previous block through a
skip-connection.
Generally, the local maxima coincide with the skip-connections while the
minima, with the two intermediary residual outputs.
It seems that the sharp decreases indicate that some of the attribute
information is discarded, possibly because it is not relevant for the
recognition output.

\begin{figure}[t]
  \includegraphics[width=0.49\textwidth]{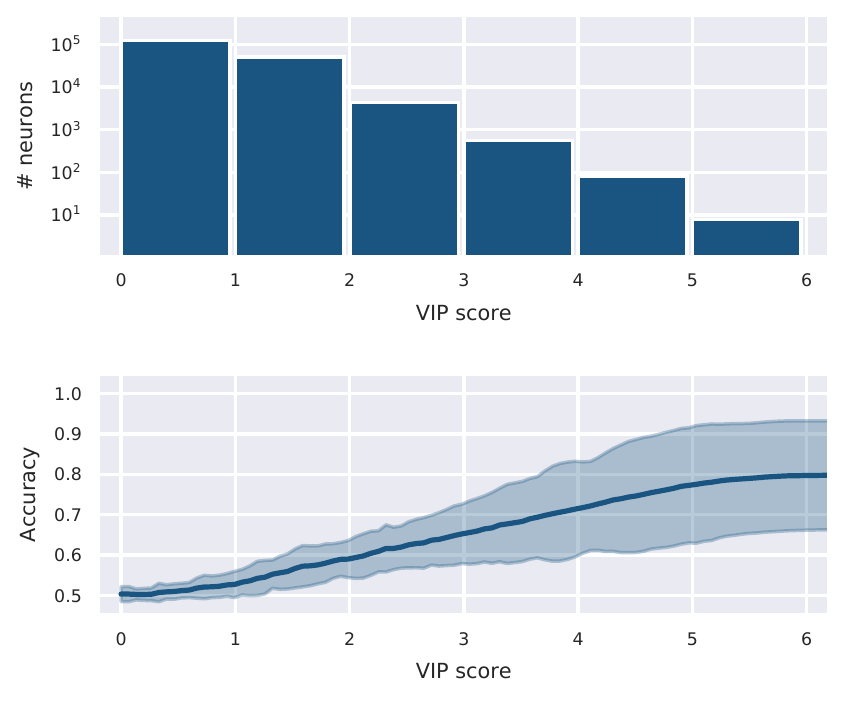} 
  \caption{
  Top: average distribution of the neural VIP score over all layers of the
  DCNN (please note the log-scale in the y-axis).
  Bottom: average accuracy using a single neuron as a feature, the shaded area
  represents two standard deviations.
    Both analysis were performed for the Male attribute.
}
  \label{fig:vip_analysis}
\end{figure}

We also analyze the performance of the neurons individually to understand how
the information is distributed inside the layers.
Fig.~\ref{fig:vip_analysis} shows the neural VIP score distribution and its
average accuracy for the Male attribute.
We record the VIP score of all neurons for each layer and report the average
distribution over all layers (notice the log-scale).
It is clear that high-scoring neurons are very scarce, suggesting that the
information for each attribute is concentrated in only a few neurons.
We then measure, for each layer, the classifier accuracy when only a single
neuron is available as input, using $50$ randomly selected neurons per layer,
sampled uniformly with respect to their VIP score.
The average performance shows, as expected, that the VIP score is a
good indicator of which neurons encodes the relevant information.

\vspace{1em}

\begin{table*}[t!]
  \centering
  \begin{tabular}{l | c c c c c c c | c c c c c c c | c}
    & \rot{Male}        & \rot{Smiling}     & \rot{Eyeglasses}  & \rot{W. Hat} & \rot{W. Lipstick}& \rot{Wavy Hair}   & \rot{H. Cheekbones}& \rot{Pointy Nose} & \rot{Big Nose}    & \rot{Blurry}      & \rot{Brown Hair}  & \rot{Oval Face}   & \rot{Narrow Eyes} & \rot{W. Necklace}& \rot{Average}     \\
    \hline
    MOON~\cite{rudd2016moon} & - & -& -& -& -& -& -& -& -& -& -& -& -& -& 90.9\\
    AttCNN~\cite{hand2018doing} & - & -& -& -& -& -& -& -& -& -& -& -& -& -& 91.0\\
    LNets + ANets~\cite{liu2015deep} & 98.0& 92.0& 99.0& 99.0& 93.0& 80.0& 87.0& 72.0& 78.0& 84.0& 80.0& 66.0& 81.0& 71.0& 87.0\\
    Multitask           & 98.2& 91.3& 99.5& 98.9& 93.7& 82.3& 86.2& 76.6& 83.9& 95.2& 86.2& 73.7& 85.6& 86.2& 90.2\\
    \hline
    Best Layer          & 97.7& 90.4& 98.3& 97.5& 92.1& 80.2& 83.6& 65.3& 72.6& 83.4& 74.0& 60.4& 71.3& 62.7& 83.1\\
    Best Filter         & 96.4& 87.9& 96.7& 95.1& 89.9& 77.3& 81.0& 62.6& 69.6& 76.8& 67.3& 59.4& 72.4& 54.8& 79.5\\
    Best Neuron         & 96.2& 83.4& 96.1& 96.3& 89.6& 70.5& 78.8& 65.3& 68.7& 69.8& 64.2& 60.3& 75.2& 54.7& 77.4\\
  \end{tabular}
  \caption{Accuracy on the CelebA test set.}
  \label{tab:comparison}
\end{table*}

{
Since the attribute information is concentrated in only a few neurons, our
training process can be simplified to use the relevant neurons directly as
input, without the need of projecting the output into a low-dimensional space.
}
The orange curve in Fig.~\ref{fig:pls_layerwise} shows the performance of the
QDA classifier using the eight highest scoring neurons of each layer.
{
For mid-level representations, the curves behave as expected, the
classifier with eight neurons as input has a comparable accuracy to the one
with the entire layer as input.
However, for most attributes, there is a large performance gap in the first
and last layers of the network.
}
We believe that the small receptive fields are the main reason of the poor
performance in the first layers, as the neurons would have to make decisions
based only on a partial input.
For the last layers, it is possible that the neurons begin to be
specialized for identification, and thus, the attribute information is spread
out among many identities and neurons.
The green curve in Fig.~\ref{fig:pls_layerwise} shows the achieved accuracy
when entire channels are selected, instead of neurons.
We select eight channels from each layer and then project these channels into
eight components with PLS.
This time, the gaps in the initial layers are reduced, supporting our
hypothesis that the shallow neurons lack the receptive field.

\begin{figure}[!t]
  \includegraphics{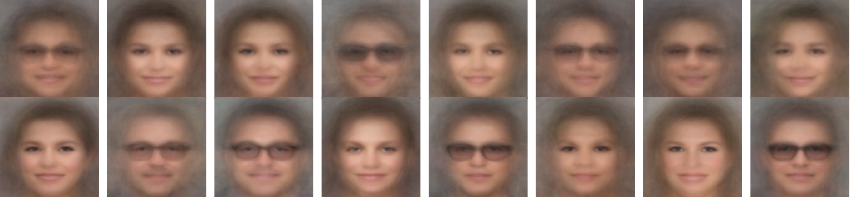} 
  \caption{
    Each column contains an average image for the highest (top) and the lowest
    (bottom) activation of a neuron. These are most discriminative neurons of
    the 34th layer, under the VIP criterion, for the Eyeglasses attribute.
  }
  \label{fig:avg_faces}
\end{figure}

While one of our main goals was to avoid the subjectivity of visualizations,
we provide a very simplistic representation of the neurons in the network to
show a flaw that is present in many other techniques.
First, we measure the activation of the top-8 neurons of an attribute, using
all images in the test split of the CelebA dataset.
Then, for each neuron, we create an average of the $100$ images that produce
the highest activations, and another average image for those with the lowest.
Fig.~\ref{fig:avg_faces} shows the obtained images for the top-8 neurons
encoding the Eyeglasses attribute.
We can notice that the attribute does not necessarily produce a high
activation in the discriminative neuron, thus, maximization of
hidden neurons does not necessarily produce good or even correct
representations of what is really encoded.

Finally, we make a quantitative comparison between deep recognition features
with state-of-the-art approaches for attribute prediction on the CelebA test
set.
Our goal here is to show that, for some of the attributes, the features of a
recognition DCNN, which were not supervised for attribute recognition, are
competitive against fully supervised deep learning methods.
We provide an additional baseline by finetunning each of our six models for
attribute prediction, by simply replacing the identification output layer for
an attribute prediction multitask layer.
Table~\ref{tab:comparison} shows the obtained performance for some attributes,
as well as the average of all attributes.
The best layer, filter and neuron were chosen based on their performance on
the validation set.
To reduce the number of candidates, the best filters and neurons were chosen
based on a list of top-8 highest VIP scores for each layer of the model.
For conciseness, we only report a few attributes, which highlight the best and
worst attributes of our analysis, respectively on the left and right halves of
the table.
The reported average is calculated over all $40$ attributes.

The average results achieved in Table~\ref{tab:comparison} show that, for each
attribute, a single neuron is less than $6$ p.p. worse than an entire
high-dimensional layer.
Selecting the best filter diminishes this gap to less than $4$ p.p.,
indicating that the attributes are captured by specific units in the network.
The small gap can be explained by lack of robustness of a single filter or
neuron, which may not be able to capture all variations of the attribute.
The left half of Table~\ref{tab:comparison} shows attributes in which 
the deep recognition features are less than $3$ p.p. worse than a fully
supervised multitask network.

\section{Conclusions and Future Works}
While very effective, deeply learned representations are difficult to
understand.
Previous attempts to understand these representations were mainly based on
subjective visualizations, which could lead to inaccurate conclusions.
Through extensive experiments, we were able to quantitatively show that many
face attributes are encoded inside a deep face recognition network.
When compared to deep techniques that were fully supervised for attribute
recognition, features from a face identification DCNN remain competitive,
demonstrating that they are able to learn these concepts implicitly.
Using our proposed method, we were able to evaluate high-dimensional layers of
a deep network and identify the key feature points where each attribute
information is encoded.
Our analysis also revealed that these semantic concepts are highly
concentrated on a few neurons, instead of being entangled in the entire layer.
Furthermore, we observed that the neural information is not necessarily
encoded with maximal activations, which was an assumption made by many of the
previous visualization techniques.

We also observed that a few similar attributes displayed vastly different
performances, perhaps due to dataset biases or failures cases of deep
representations, which we intend to further investigate in future works.
Our analysis was focused on the ResNet50 architecture, revealing some
interesting properties of the residual blocks.
{
We believe that applying our proposed framework to different networks,
could also expose new interesting properties and behaviors which may aid
other researches to develop better architectures.
}



\section*{Acknowledgments}

The authors would like to thank the National Council for Scientific and
Technological Development -- CNPq (Grants~438629/2018-3 and~309953/2019-7),
the Minas Gerais Research Foundation -- FAPEMIG (Grants~APQ-00567-14
and~PPM-00540-17) and the Coordination for the Improvement of Higher Education
Personnel -- CAPES (DeepEyes Project and Finance Code 001).

\balance
\bibliographystyle{ieee}
\bibliography{refs}

\begin{thebibliography}{10}\itemsep=-1pt

\bibitem{abdi2010partial}
H.~Abdi.
\newblock Partial least squares regression and projection on latent structure
  regression (pls regression).
\newblock {\em Wiley interdisciplinary reviews: computational statistics},
  2(1):97--106, 2010.

\bibitem{cao2018vggface2}
Q.~Cao, L.~Shen, W.~Xie, O.~M. Parkhi, and A.~Zisserman.
\newblock Vggface2: A dataset for recognising faces across pose and age.
\newblock In {\em 2018 13th IEEE International Conference on Automatic Face \&
  Gesture Recognition (FG 2018)}, pages 67--74. IEEE, 2018.

\bibitem{deng2019arcface}
J.~Deng, J.~Guo, N.~Xue, and S.~Zafeiriou.
\newblock Arcface: Additive angular margin loss for deep face recognition.
\newblock In {\em Proceedings of the IEEE Conference on Computer Vision and
  Pattern Recognition}, pages 4690--4699, 2019.

\bibitem{deng2017marginal}
J.~Deng, Y.~Zhou, and S.~Zafeiriou.
\newblock Marginal loss for deep face recognition.
\newblock In {\em Proceedings of the IEEE Conference on Computer Vision and
  Pattern Recognition Workshops}, pages 60--68, 2017.

\bibitem{dosovitskiy2016inverting}
A.~Dosovitskiy and T.~Brox.
\newblock Inverting visual representations with convolutional networks.
\newblock In {\em Proceedings of the IEEE Conference on Computer Vision and
  Pattern Recognition}, pages 4829--4837, 2016.

\bibitem{ferrari2019discovering}
C.~Ferrari, S.~Berretti, and A.~Del~Bimbo.
\newblock Discovering identity specific activation patterns in deep descriptors
  for template based face recognition.
\newblock In {\em 2019 14th IEEE International Conference on Automatic Face \&
  Gesture Recognition (FG 2019)}, pages 1--5. IEEE, 2019.

\bibitem{geladi1986partial}
P.~Geladi and B.~R. Kowalski.
\newblock Partial least-squares regression: a tutorial.
\newblock {\em Analytica chimica acta}, 185:1--17, 1986.

\bibitem{hand2018doing}
E.~M. Hand, C.~Castillo, and R.~Chellappa.
\newblock Doing the best we can with what we have: Multi-label balancing with
  selective learning for attribute prediction.
\newblock In {\em Thirty-Second AAAI Conference on Artificial Intelligence},
  2018.

\bibitem{he2016deep}
K.~He, X.~Zhang, S.~Ren, and J.~Sun.
\newblock Deep residual learning for image recognition.
\newblock In {\em Proceedings of the IEEE conference on computer vision and
  pattern recognition}, pages 770--778, 2016.

\bibitem{huang2008labeled}
G.~B. Huang, M.~Mattar, T.~Berg, and E.~Learned-Miller.
\newblock Labeled faces in the wild: A database forstudying face recognition in
  unconstrained environments.
\newblock 2008.

\bibitem{jordao2018latent}
A.~Jordao, R.~Kloss, and W.~R. Schwartz.
\newblock Latent hypernet: Exploring the layers of convolutional neural
  networks.
\newblock In {\em 2018 International Joint Conference on Neural Networks
  (IJCNN)}, pages 1--7. IEEE, 2018.

\bibitem{jordao2018pruning}
A.~Jordao, F.~Yamada, and W.~R. Schwartz.
\newblock Pruning deep neural networks using partial least squares.
\newblock {\em arXiv preprint arXiv:1810.07610}, 2018.

\bibitem{kemelmacher2016megaface}
I.~Kemelmacher-Shlizerman, S.~M. Seitz, D.~Miller, and E.~Brossard.
\newblock The megaface benchmark: 1 million faces for recognition at scale.
\newblock In {\em Proceedings of the IEEE Conference on Computer Vision and
  Pattern Recognition}, pages 4873--4882, 2016.

\bibitem{kong2016hypernet}
T.~Kong, A.~Yao, Y.~Chen, and F.~Sun.
\newblock Hypernet: Towards accurate region proposal generation and joint
  object detection.
\newblock In {\em Proceedings of the IEEE conference on computer vision and
  pattern recognition}, pages 845--853, 2016.

\bibitem{krizhevsky2012imagenet}
A.~Krizhevsky, I.~Sutskever, and G.~E. Hinton.
\newblock Imagenet classification with deep convolutional neural networks.
\newblock In {\em Advances in neural information processing systems}, pages
  1097--1105, 2012.

\bibitem{liu2015deep}
Z.~Liu, P.~Luo, X.~Wang, and X.~Tang.
\newblock Deep learning face attributes in the wild.
\newblock In {\em Proceedings of the IEEE international conference on computer
  vision}, pages 3730--3738, 2015.

\bibitem{mahendran2016visualizing}
A.~Mahendran and A.~Vedaldi.
\newblock Visualizing deep convolutional neural networks using natural
  pre-images.
\newblock {\em International Journal of Computer Vision}, 120(3):233--255,
  2016.

\bibitem{mehmood2012review}
T.~Mehmood, K.~H. Liland, L.~Snipen, and S.~S{\ae}b{\o}.
\newblock A review of variable selection methods in partial least squares
  regression.
\newblock {\em Chemometrics and Intelligent Laboratory Systems}, 118:62--69,
  2012.

\bibitem{nguyen2016multifaceted}
A.~Nguyen, J.~Yosinski, and J.~Clune.
\newblock Multifaceted feature visualization: Uncovering the different types of
  features learned by each neuron in deep neural networks.
\newblock {\em arXiv preprint arXiv:1602.03616}, 2016.

\bibitem{ranjan2017l2}
R.~Ranjan, C.~D. Castillo, and R.~Chellappa.
\newblock L2-constrained softmax loss for discriminative face verification.
\newblock {\em arXiv preprint arXiv:1703.09507}, 2017.

\bibitem{ranjan2017all}
R.~Ranjan, S.~Sankaranarayanan, C.~D. Castillo, and R.~Chellappa.
\newblock An all-in-one convolutional neural network for face analysis.
\newblock In {\em 2017 12th IEEE International Conference on Automatic Face \&
  Gesture Recognition (FG 2017)}, pages 17--24. IEEE, 2017.

\bibitem{rudd2016moon}
E.~M. Rudd, M.~G{\"u}nther, and T.~E. Boult.
\newblock Moon: A mixed objective optimization network for the recognition of
  facial attributes.
\newblock In {\em European Conference on Computer Vision}, pages 19--35.
  Springer, 2016.

\bibitem{schroff2015facenet}
F.~Schroff, D.~Kalenichenko, and J.~Philbin.
\newblock Facenet: A unified embedding for face recognition and clustering.
\newblock In {\em Proceedings of the IEEE conference on computer vision and
  pattern recognition}, pages 815--823, 2015.

\bibitem{schwartz2009human}
W.~R. Schwartz, A.~Kembhavi, D.~Harwood, and L.~S. Davis.
\newblock Human detection using partial least squares analysis.
\newblock In {\em 2009 IEEE 12th International Conference on Computer Vision
  (ICCV)}, pages 24--31. IEEE, 2009.

\bibitem{selvaraju2017grad}
R.~R. Selvaraju, M.~Cogswell, A.~Das, R.~Vedantam, D.~Parikh, and D.~Batra.
\newblock Grad-cam: Visual explanations from deep networks via gradient-based
  localization.
\newblock In {\em Proceedings of the IEEE International Conference on Computer
  Vision}, pages 618--626, 2017.

\bibitem{shafiee2018efficient}
M.~S. Shafiee, M.~J. Shafiee, and A.~Wong.
\newblock Efficient inference on deep neural networks by dynamic
  representations and decision gates.
\newblock {\em arXiv preprint arXiv:1811.01476}, 2018.

\bibitem{simonyan2013deep}
K.~Simonyan, A.~Vedaldi, and A.~Zisserman.
\newblock Deep inside convolutional networks: Visualising image classification
  models and saliency maps.
\newblock {\em arXiv preprint arXiv:1312.6034}, 2013.

\bibitem{sun2014deep}
Y.~Sun, Y.~Chen, X.~Wang, and X.~Tang.
\newblock Deep learning face representation by joint
  identification-verification.
\newblock In {\em Advances in neural information processing systems}, pages
  1988--1996, 2014.

\bibitem{sun2015deeply}
Y.~Sun, X.~Wang, and X.~Tang.
\newblock Deeply learned face representations are sparse, selective, and
  robust.
\newblock In {\em Proceedings of the IEEE conference on computer vision and
  pattern recognition}, pages 2892--2900, 2015.

\bibitem{szegedy2013intriguing}
C.~Szegedy, W.~Zaremba, I.~Sutskever, J.~Bruna, D.~Erhan, I.~Goodfellow, and
  R.~Fergus.
\newblock Intriguing properties of neural networks.
\newblock {\em arXiv preprint arXiv:1312.6199}, 2013.

\bibitem{yosinski2015understanding}
J.~Yosinski, J.~Clune, A.~Nguyen, T.~Fuchs, and H.~Lipson.
\newblock Understanding neural networks through deep visualization.
\newblock {\em arXiv preprint arXiv:1506.06579}, 2015.

\bibitem{zeiler2014visualizing}
M.~D. Zeiler and R.~Fergus.
\newblock Visualizing and understanding convolutional networks.
\newblock In {\em European conference on computer vision}, pages 818--833.
  Springer, 2014.

\bibitem{zhao2019regularface}
K.~Zhao, J.~Xu, and M.-M. Cheng.
\newblock Regularface: Deep face recognition via exclusive regularization.
\newblock In {\em Proceedings of the IEEE Conference on Computer Vision and
  Pattern Recognition}, pages 1136--1144, 2019.

\bibitem{zhong2019exploring}
Y.~Zhong and W.~Deng.
\newblock Exploring features and attributes in deep face recognition using
  visualization techniques.
\newblock In {\em 2019 14th IEEE International Conference on Automatic Face \&
  Gesture Recognition (FG 2019)}, pages 1--8. IEEE, 2019.

\bibitem{zhou2014object}
B.~Zhou, A.~Khosla, A.~Lapedriza, A.~Oliva, and A.~Torralba.
\newblock Object detectors emerge in deep scene cnns.
\newblock {\em arXiv preprint arXiv:1412.6856}, 2014.

\bibitem{zhou2016learning}
B.~Zhou, A.~Khosla, A.~Lapedriza, A.~Oliva, and A.~Torralba.
\newblock Learning deep features for discriminative localization.
\newblock In {\em Proceedings of the IEEE conference on computer vision and
  pattern recognition}, pages 2921--2929, 2016.

\end{thebibliography}


\begin{thebibliography}{1}\itemsep=-1pt

\bibitem{ranjan2017l2}
R.~Ranjan, C.~D. Castillo, and R.~Chellappa.
\newblock L2-constrained softmax loss for discriminative face verification.
\newblock {\em arXiv preprint arXiv:1703.09507}, 2017.

\end{thebibliography}

\end{document}